\documentclass[letterpaper, 10 pt, conference]{ieeeconf}  

\IEEEoverridecommandlockouts                              

\usepackage{geometry}

\usepackage{colortbl}
\usepackage{graphics} 
\usepackage{epsfig} 
\usepackage{mathptmx} 
\usepackage{times} 
\usepackage{amsmath} 
\usepackage{amssymb}  
\usepackage{dashrule}
\usepackage{booktabs}
\usepackage{multirow}
\usepackage{float}
\usepackage{tablefootnote}
\usepackage{placeins}
\usepackage{colortbl}    
\usepackage{xcolor}      
\usepackage{graphicx}    
\usepackage{subfig}

\newcommand{\pf}[1]{\cellcolor{red!40}#1}
\newcommand{\ps}[1]{\cellcolor{orange!50}#1}
\newcommand{\pt}[1]{\cellcolor{yellow!50}#1}

\title{\LARGE \bf
R3GS: Gaussian Splatting for Robust Reconstruction and Relocalization in Unconstrained Image Collections
}

\author{Xu Yan$^{1}$, Zhaohui Wang$^{2}$, Rong Wei$^{3}$, Jingbo Yu$^{4}$, Dong Li$^{5}$ and Xiangde Liu$^{1,\star}$
\thanks{$^{1}$All authors are with the Beijing Digital Native Digital City Research Center,
        Zhiyuan Building, No. 150 Chengfu Road, Haidian District, Beijing, China
        {\tt\small yanxu@bdnrc.org.cn}}%
\thanks{$^\star$ Xiangde Liu is the corresponding author.}
}

\begin{document}

\maketitle
\thispagestyle{empty}
\pagestyle{empty}

\begin{abstract}
3D Gaussian Splatting (3DGS) \cite{kerbl3DGaussianSplatting2023} has demonstrated not only comparable rendering quality to existing novel view synthesis (NVS) methods but also superior rendering speed, enhancing its potential for real-world applications. However, when applied to outdoor, unconstrained image collections, several challenges arise, such as transient objects, varying scene appearances, and sky regions, which affect rendering quality and visual relocalization accuracy.
To handle these issues, we propose R3GS, a robust reconstruction and relocalization framework tailored for unconstrained datasets. Our method uses a hybrid representation during training. Each anchor combines a global feature from a convolutional neural network (CNN) with a local feature encoded by the multiresolution hash grids \cite{mullerInstantNeuralGraphics2022}. Subsequently, several shallow multi-layer perceptrons (MLPs) predict the attributes of each Gaussians, including color, opacity, and covariance.
To mitigate the adverse effects of transient objects on the reconstruction process, we fine-tune a lightweight human detection network. Once fine-tuned, this network generates a visibility map that efficiently generalizes to other transient objects (such as posters, banners, and cars) with minimal need for further adaptation. Additionally, to address the challenges posed by sky regions in outdoor scenes, we propose an effective sky-handling technique that incorporates a depth prior as a constraint. This allows the infinitely distant sky to be represented on the surface of a large-radius sky sphere, significantly reducing floaters caused by errors in sky reconstruction.
Furthermore, we introduce a novel relocalization method that remains robust to changes in lighting conditions while estimating the camera pose of a given image within the reconstructed 3DGS scene.
As a result, R3GS significantly enhances rendering fidelity, improves both training and rendering efficiency, and reduces storage requirements. Our method achieves state-of-the-art performance compared to baseline methods on in-the-wild datasets. The code will be made open-source following the acceptance of the paper.
\end{abstract}

\begin{figure}[!t]
	\centering
	\includegraphics[width=0.5\textwidth]{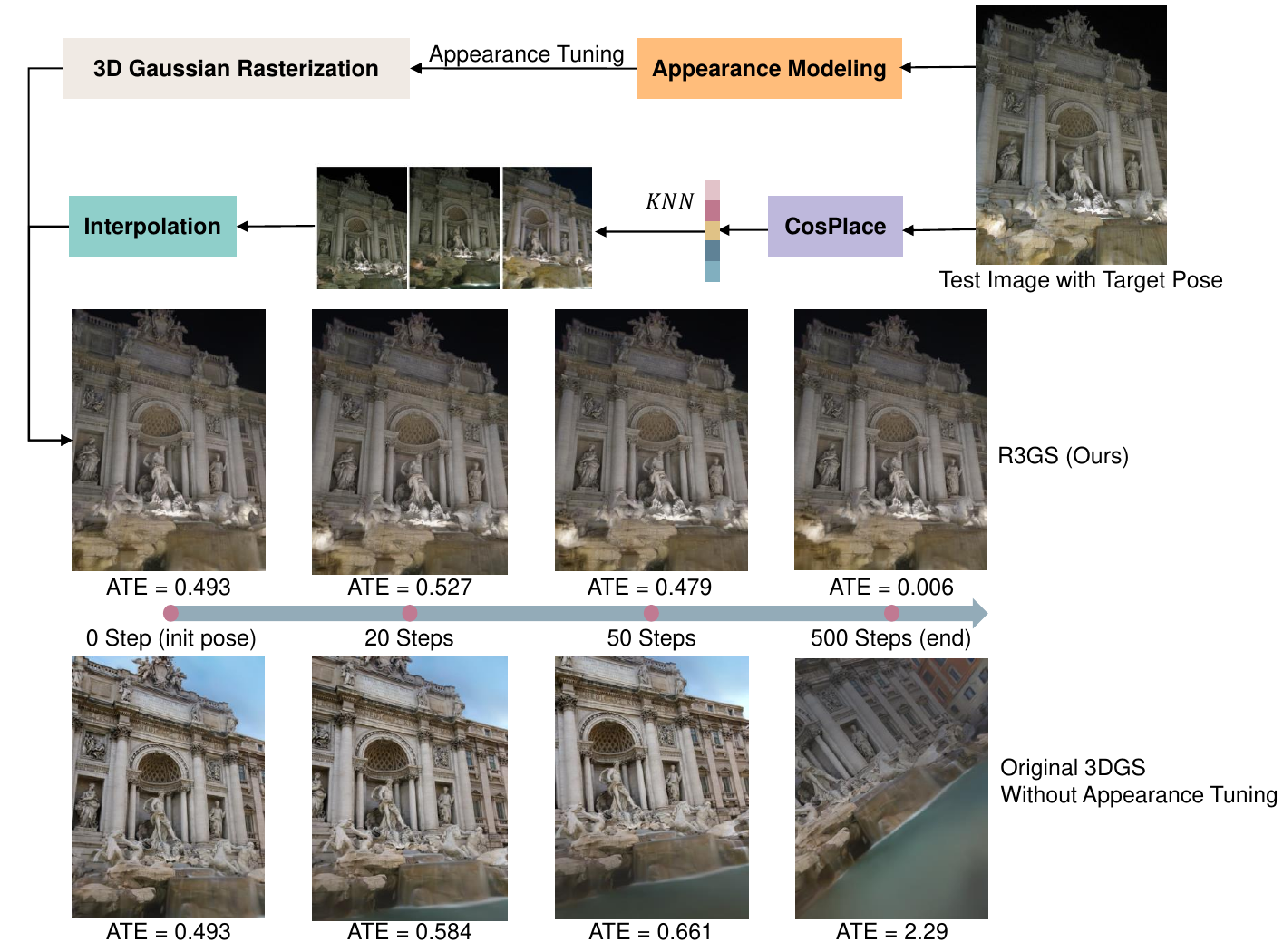}
	\caption{Given an unconstrained image collection as input, R3GS achieves high relocalization accuracy through appearance tuning, significantly outperforming 3DGS in terms of Absolute Trajectory Error (ATE).}
 \vspace{-6mm}
	\label{fig:relocalization}
\end{figure}
\section{INTRODUCTION}

Novel view synthesis (NVS) has received significant attention due to its vast potential applications in many fields, such as virtual reality (VR) and 3D environment simulations in robotics. NVS aims to learn a scene representation from a set of 2D images and render novel views for unseen perspectives.
In order to adress the task of NVS, Neural Radiance Fields (NeRF) \cite{mildenhallNeRFRepresentingScenes2020} is widely recognized for its ability to generate high-quality, photorealistic scene reconstructions by modeling volumetric radiance fields. However, a critical limitation is its slow rendering speed, particularly in interactive settings. This slowness stems from the need to query the neural network multiple times for each pixel, across various viewing angles. Specifically, for each pixel, hundreds of points along a ray are sampled, which results in an intensive computational process that hampers real-time performance and user experience.

Conversely, 3D Gaussian Splatting (3DGS) \cite{kerbl3DGaussianSplatting2023} has emerged as a compelling alternative to NeRF, offering significant improvements in both training and rendering speed while maintaining comparable quality in novel view synthesis. By representing the scene explicitly with 3D Gaussians, 3DGS employs a differentiable Gaussian rasterization approach, enabling real-time rendering with high fidelity.

In real-world outdoor scenes, appearance variations are highly significant, and transient objects such as vehicles and pedestrians can degrade the quality of scene representation. Additionally, effective sky modeling plays a crucial role in the visual fidelity of the reconstructed scene. NeRF-W \cite{martin2021nerf} was the first to address these challenges by introducing generative latent optimization, which assigns per-image latent codes to optimize for varying appearances and transient objects. Building on NeRF-W, subsequent NeRF-based methods \cite{chen2022hallucinated,yang2023cross,garbin2021fastnerf,muller2022instant} introduced more sophisticated networks to improve transient object prediction and enhance the extraction of appearance features from input images.

However, despite the innovative solutions for handling these challenges, NeRF-based methods are still constrained by slow training and rendering speeds, which limits their practical application. In contrast, 3DGS offers a promising alternative by leveraging explicit scene representation, significantly improving efficiency.

In this work, we introduce R3GS, a novel framework that integrates the strengths of NeRF's implicit representation and 3DGS's explicit representation. The primary contributions of this paper can be summarized as follows:
\begin{itemize}
\item \textbf{Hybrid Representation}: R3GS integrates a global feature from a convolutional neural network (CNN) with locally encoded features using multiresolution hash grids. This hybrid representation allows for precise attribute prediction, including color, opacity, and covariance, via shallow multi-layer perceptrons (MLPs).

\begin{figure*}[t!]
	\centering
	\includegraphics[width=0.8\textwidth]{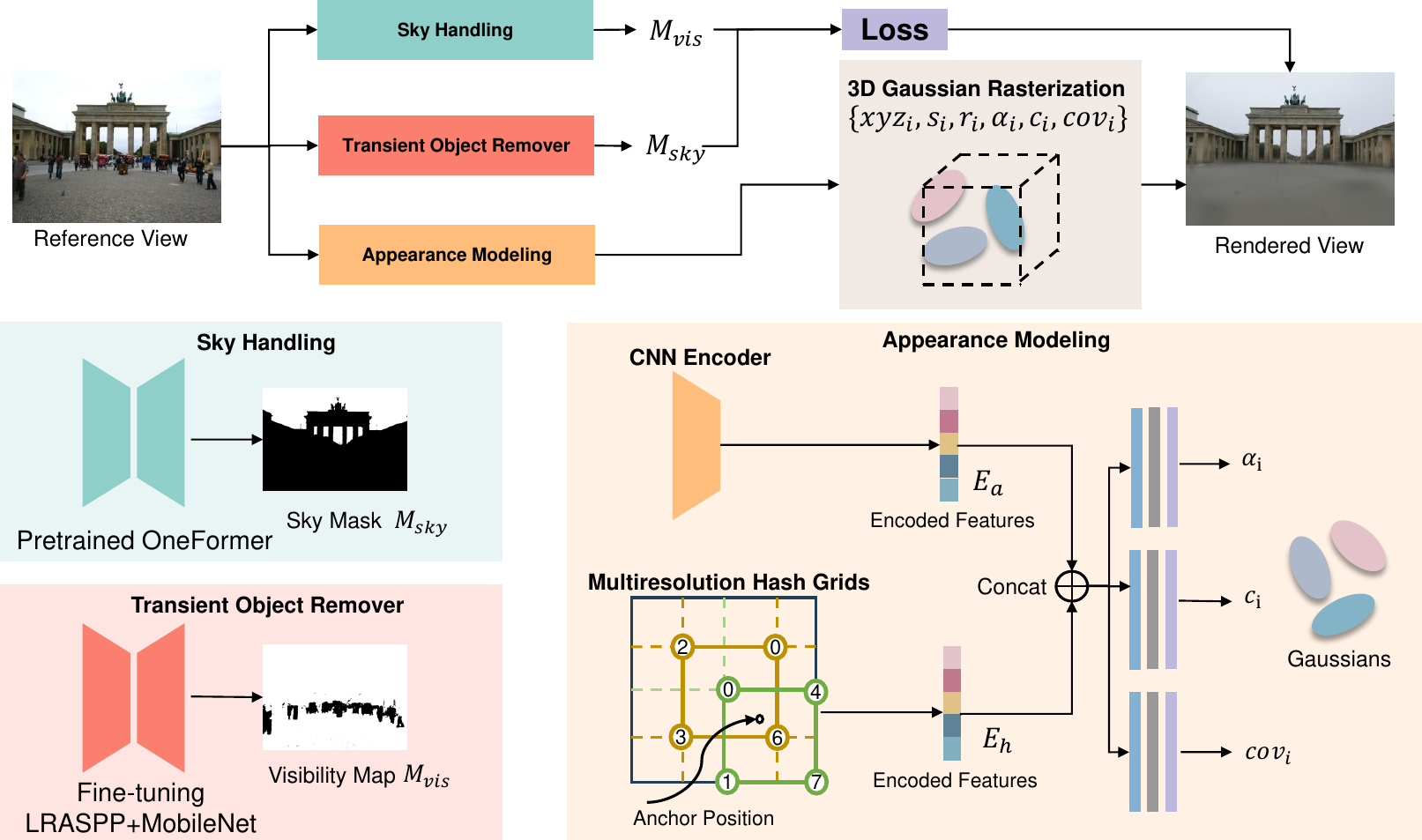}
	\caption{Proposed R3GS model architecture-We begin with a reference view $I$ and its camera pose $T$, anchor are initialized from sfm \cite{schonberger2016structure} points and sampled points in a sky sphere. Per-anchor local feature queried from Multiresolution Hash Grid and per-image global feature extracted from $I$ by utilizing CNN network are concatenated to fed into 3 individual MLPs to predict the three attributes (color, opacity, covariance) of the neural Gaussians respectively. The generated non-trivial neural Gaussians ($opacity > 0$) are rasterized following original 3DGS \cite{kerbl3DGaussianSplatting2023}. A pretrained encoder-decoder network capable of detecting human figures is used to generate visibility maps. After 5000 iterations of fine-tuning, the network can converge and detect other transient objects such as vehicles. The sky mask $M_{sky}$ generated by OneFormer \cite{jain2023oneformer} is utilized in the optimization process, where depth regularization is applied to constrain the Gaussians to the sky sphere.}
	\label{fig:pipeline}
\end{figure*}

\item \textbf{Transient Object Remover}: We fine-tune a lightweight human detection network, which generates a visibility map. This map generalizes efficiently to other transient objects, such as posters, banners, and cars, reducing the negative impact of transient elements on scene reconstruction.

\item \textbf{Sky Handling Technique}: To handle sky regions, we propose a novel technique incorporating depth prior as a constraint. This allows the sky to be represented on a large-radius sky sphere, effectively mitigating rendering errors like floaters.

\item \textbf{Robust Relocalization}: R3GS introduces a novel relocalization method that remains robust to lighting variations, ensuring accurate camera pose estimation in the reconstructed 3DGS scenes.

\end{itemize}
\section{RELATED WORK}
\subsection{3D Gaussian Splatting (3DGS)}

3DGS \cite{kerbl3DGaussianSplatting2023} often leverages millions of learnable 3D Gaussians to explicitly represent scenes, utilizing a CUDA-accelerated renderer to efficiently synthesize novel views. This approach not only surpasses NeRF in both rendering speed and visual fidelity but also introduces a number of challenges. The reliance on millions of Gaussians significantly increases CUDA memory consumption, leading to higher memory demands for 3DGS-based scenes. Additionally, the specialized rendering pipeline poses complexities in optimizing the framework, making it less adaptable compared to other approaches.

Recent advances such as Scaffold-GS \cite{lu2024scaffold} aim to mitigate CUDA memory overhead by integrating multi-layer perceptrons (MLPs) with 3DGS. Meanwhile, techniques like VastGaussians \cite{lin2024vastgaussian}, PyGS \cite{wang2024pygs}, and GS-LRM \cite{zhang2024gs} explore the application of 3DGS for large-scale urban scene reconstruction. Other research efforts, such as SplaTAM \cite{keetha2024splatam}, RTG-SLAM \cite{peng2024rtgslam} and MonoGS \cite{matsuki2024gaussian}, incorporate 3DGS into simultaneous localization and mapping (SLAM) frameworks, while DrivingGaussian \cite{zhou2024drivinggaussian}, GaussianBEV \cite{chabot2024gaussianbev} and GaussianFormer \cite{huang2024gaussianformer} investigates the use of 3DGS in autonomous driving scenarios. Despite these promising developments, unique challenges persist in extending 3DGS to outdoor, unconstrained datasets, particularly in terms of scalability and robustness.

\subsection{3DGS in outdoor scenes}
In unconstrained image collections, scenes often exhibit significant variations in appearance, transient objects, and sky conditions, presenting unique challenges for 3D Gaussian Splatting (3DGS). To address appearance variability, Wild-GS \cite{xu2024wild} employs a triplane projection technique, utilizing three encoded feature planes to represent spherical harmonic color coefficients and a self-supervised U-Net to model transient object dynamics. Gaussian in the Wild \cite{zhang2024gaussian} improves flexibility by assigning trainable attributes to each Gaussian, discarding the traditional spherical harmonic color representation in favor of a novel renderer capable of handling dynamic appearances. WildGaussians \cite{kulhanek2024wildgaussians} identifies performance degradation in sky rendering due to poor initialization of the point cloud, addressing this by sampling from a 3D sphere to create the initial Gaussian set. Furthermore, Splatfacto-W \cite{xu2024splatfacto} introduces a separation of foreground and background, rendering these elements independently before compositing them into the final scene, improving handling of sky and foreground variability.

While these methods successfully mitigate several challenges in handling unconstrained image collections, they significantly compromise the rendering and training efficiency of the original 3DGS framework, and impose considerable storage requirements. To overcome these limitations, we propose R3GS, a method that effectively handles appearance changes, transient objects, and sky variations while preserving the fast rendering and training speeds of original 3DGS.

\section{METHOD}
\subsection{Preliminaries}
3D Gaussian Splatting (3DGS) \cite{kerbl3DGaussianSplatting2023} uses explicit Gaussians to model scenes, each with learnable attributes: opacity ($\alpha$), covariance matrix ($\Sigma$), position ($\mu$), and color ($c$). The influence of each Gaussian is determined by the 3D covariance matrix $\Sigma$ and position $\mu$ in world coordinates:

\begin{equation}
    G(x-\mu,\Sigma) = e^{-\frac{1}{2}(x-\mu)^T\Sigma^{-1}(x-\mu)}
\end{equation}

To simplify optimization, 3DGS decomposes the covariance matrix into a rotation matrix $R$ and a scaling matrix $S$. To reduce memory usage, $R$ is represented as a quaternion, and $S$ is expressed as a 3D vector:

\begin{equation}
    \Sigma = RS S^T R^T
\end{equation}

In contrast to NeRF, which traces rays from the camera, 3DGS projects each Gaussian directly onto the pixel plane. The 2D covariance matrix $\Sigma'$ on the pixel plane is derived via the viewing transformation $W$ and projection matrix $J$:

\begin{equation}
    \Sigma' = J W \Sigma W^T J^T
\end{equation}

Pixel colors $C(p)$ are calculated using $\alpha$-blending \cite{max1995optical}, \cite{tagliasacchi2022volume}:

\begin{equation}
    \alpha_i = 1 - \exp(-\sigma_i \delta_i)
\end{equation}

\begin{equation}
    C(p) = \sum_{i \in N} c_i \alpha_i \prod_{j=1}^{i-1}(1 - \alpha_j)
\end{equation}

Where $N$ represents points overlapping pixel $p$, and $C(p)$ is the resulting color from Gaussian splatting. The final image is compared against the ground truth to learn Gaussian properties and adjust their quantity.

\subsection{Appearance Modeling} To enable 3DGS to achieve robust reconstruction under varying appearance inputs, we introduce a CNN based network, similarly to previous method \cite{chen2022hallucinated}, to extract per-image feature to represent the appearance of the image. However, this golobal feature is insufficient to learn the local appearance variances in an image. Thus, we employ Multiresolution Hash Grid \cite{mullerInstantNeuralGraphics2022} for acquiring position aware per-Gaussian feature with spatial consistency. As shown in Fig. \ref{fig:pipeline} we follow the architecture of Scaffold-GS \cite{lu2024scaffold}, during training, the global per-image feature is concatenated with the local per-anchor feature, and the concatenated features are used as input to multiple MLPs for predicting the attributes of neural Gaussians, including opacities, colors, and covariances. Alternative to the Scaffold-GS, in R3GS the view dependent input such as viewing direction and viewing distance are removed, and making it possible to export explicit 3DGS file and enable high rendering speed.

\subsection{Transient Object Remove by Visibility Map}
In concurrent work, visibility maps play a crucial role \cite{chen2022hallucinated, yang2023cross, zhang2024gaussian, dahmani2024swagsplattingwildimages}, particularly because outdoor scenes often contain numerous transient objects, such as pedestrians, vehicles, and street lights. These transient objects are not part of the static scene to be reconstructed, and introduce artifacts to the output. Previous approaches have often employed self-supervised methods to train encoder-decoder networks \cite{yang2023cross}, but training a full network significantly impacts the reconstruction speed. Additionally, networks relying solely on self-supervision struggle to converge quickly. In our work, we utilize a pre-training-based self-supervision approach: a pre-trained model based on LRASPP-MobileNet \cite{howard2019searching}, originally trained for human detection on the COCO dataset \cite{coco_dataset}. Although LRASPP-MobileNet is primarily designed for human detection, after fine-tuning for 20,000-25,000 iterations, the network demonstrates satisfactory segmentation performance on other transient objects, such as posters, banners, and cars. Using this fine-tuned model, visibility maps are generated to effectively remove transient objects. Furthermore, since the sky region experiences frequent appearance changes, such as the presence of white clouds, we apply the visibility map only to the foreground, ensuring that elements like clouds are not misclassified as transient objects.

\begin{table*}[]
\caption{Quantitative experimental results of existing methods }
\scriptsize
\centering
\begin{tabular}{lccccccccccc} 
\toprule
\multicolumn{1}{l}{\multirow{2}{*}{Method}} & \multicolumn{3}{c}{Brandenburg Gate} & \multicolumn{3}{c}{Sacre Coeur} & \multicolumn{3}{c}{Trevi Fountain} & \multicolumn{2}{c}{Efficiency}  \\ 
\cmidrule(lr){2-4}\cmidrule(lr){5-7}\cmidrule(lr){8-10}\cmidrule(lr){11-12}
\multicolumn{1}{c}{}                        & PSNR↑  & SSIM↑   & LPIPS↓   & PSNR↑  & SSIM↑   & LPIPS↓     & PSNR↑  & SSIM↑   & LPIPS↓       & Training↓ & FPS↑                   \\ 
NeRF-W \cite{martin2021nerf}    & 24.17 & 0.8905 & 0.1670 & 19.20 & 0.8076 & 0.1915 & 18.97    & 0.6984 & 0.2652 & $>1$     & 0.026      \\ 
Ha-NeRF \cite{chen2022hallucinated}      & 24.04 & 0.8873 & 0.1391 & 20.02 & 0.8012 & 0.1710 & 20.18    & 0.6908 & 0.2225 & $>1$      & 0.024     \\ 
CR-NeRF \cite{yang2023cross}     & \pt{26.53} & 0.9003 & \ps{0.1060} & \pt{22.07} & 0.8233 & \pt{0.1520} & 21.48    & 0.7117 & \pt{0.2069} &  $>1$ & 0.022\\
3DGS \cite{kerbl3DGaussianSplatting2023}        & 19.63 & 0.8817 & 0.1378 & 17.95 & \pt{0.8455} & 0.1633 & 17.23    & 0.6963 & 0.2815 & \pf{0.06}       & \ps{289}  \\
WildGaussians \cite{kulhanek2024wildgaussians} & 25.97 & \pt{0.9182} & 0.1521 & 21.35 & 0.8425 & 0.2180 & \pt{22.95}    & \pt{0.7671} & 0.2276 & 0.40    & - \\ 
GS-W\textsuperscript{\textdagger} \cite{zhang2024gaussian}   & \pf{28.39} & \pf{0.9306} & \pf{0.0833} & \ps{23.078} & \ps{0.8626} &  \ps{0.1260} & \ps{23.14}    & \ps{0.7752} & \ps{0.1640} & \ps{0.18}  &  61 \\
\hline
\rule{0pt}{2ex} R3GS (Ours)   & \ps{27.02} & \ps{0.9240} & \pt{0.1317} & \pf{25.14} & \pf{0.9111} & \pf{0.1201} & \pf{24.45}    & \pf{0.8540} & \pf{0.1950}   & \pt{0.31} & \pt{239}   \\
\bottomrule
\multicolumn{4}{l}{\textdagger GS-W need downsample input by 2 while others don't need}\\ 
\multicolumn{12}{l}{The \colorbox{red!40}{first}, \colorbox{orange!50}{second}, and \colorbox{yellow!50}{third} best-performing methods are highlighted. We significantly outperform all baseline methods and maintain fast rendering times.}\\
\multicolumn{12}{l}{Training↓ means the time (hours) costing by 10000 iterations. Rendering FPS↑ are calculated without any caching}\\
\end{tabular}
\label{tab:1}
\end{table*}
\subsection{Sky Handling}
\begin{figure*} [!htbp]
	\centering
	\subfloat[\label{fig:a}]{
		\includegraphics[scale=0.1]{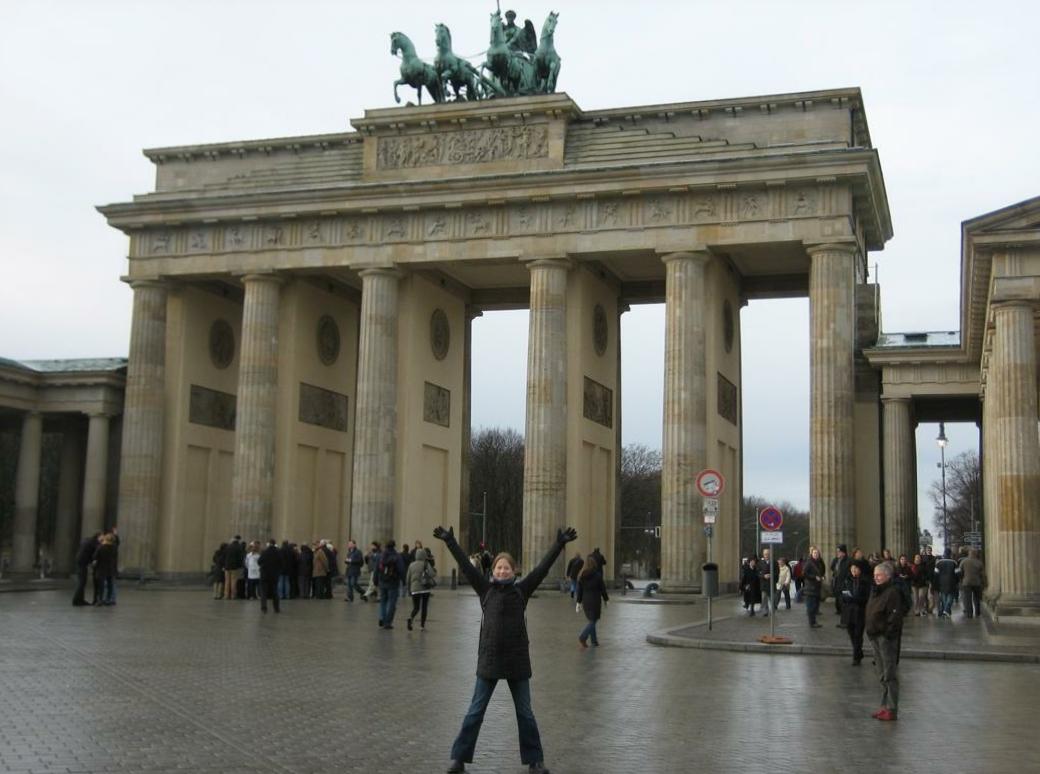}}
	\subfloat[\label{fig:b}]{
		\includegraphics[scale=0.1]{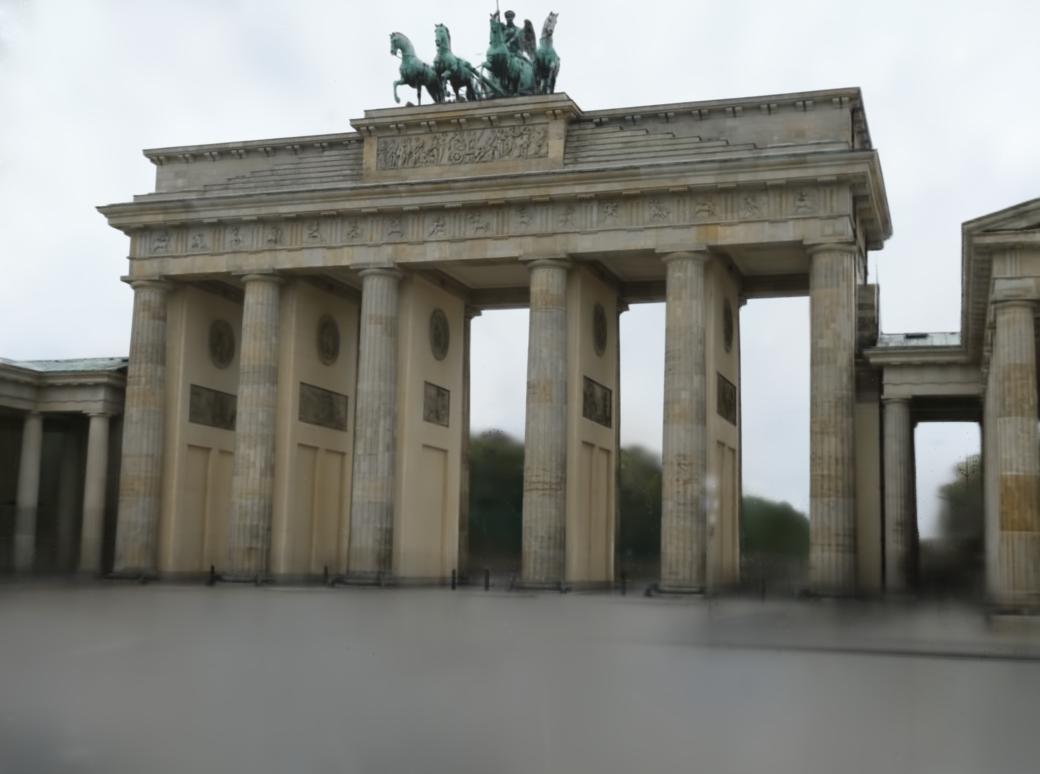}}
  	\subfloat[\label{fig:c}]{
		\includegraphics[scale=0.2]{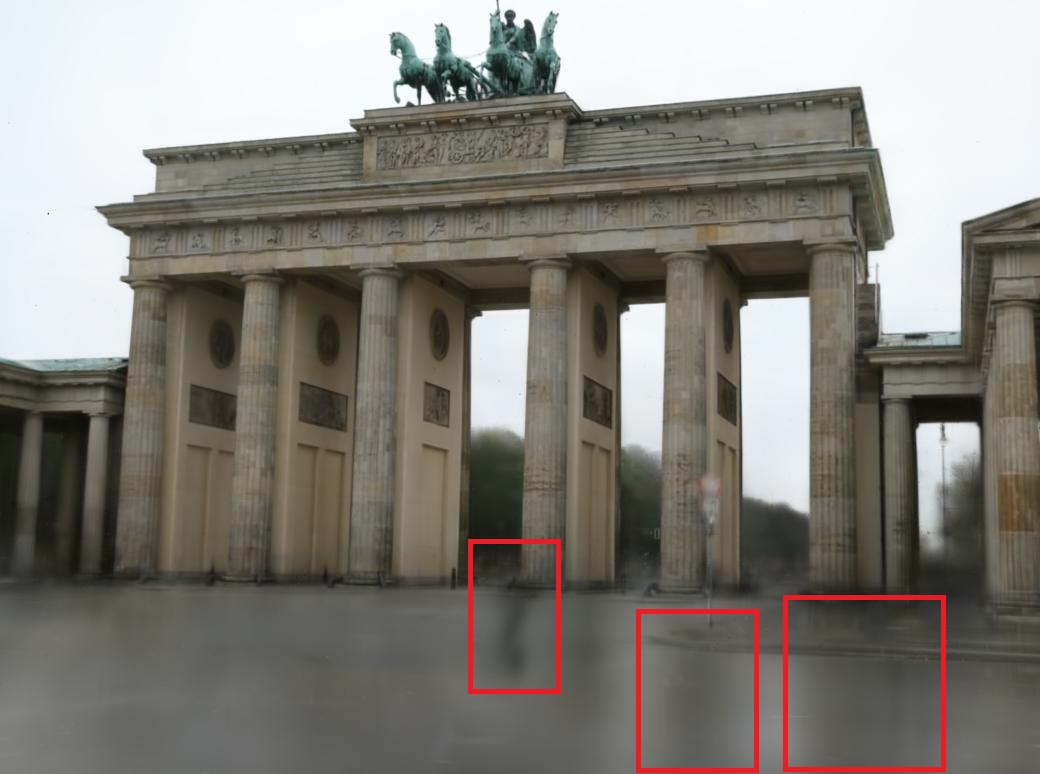}}
	\subfloat[\label{fig:d}]{
		\includegraphics[scale=0.1]{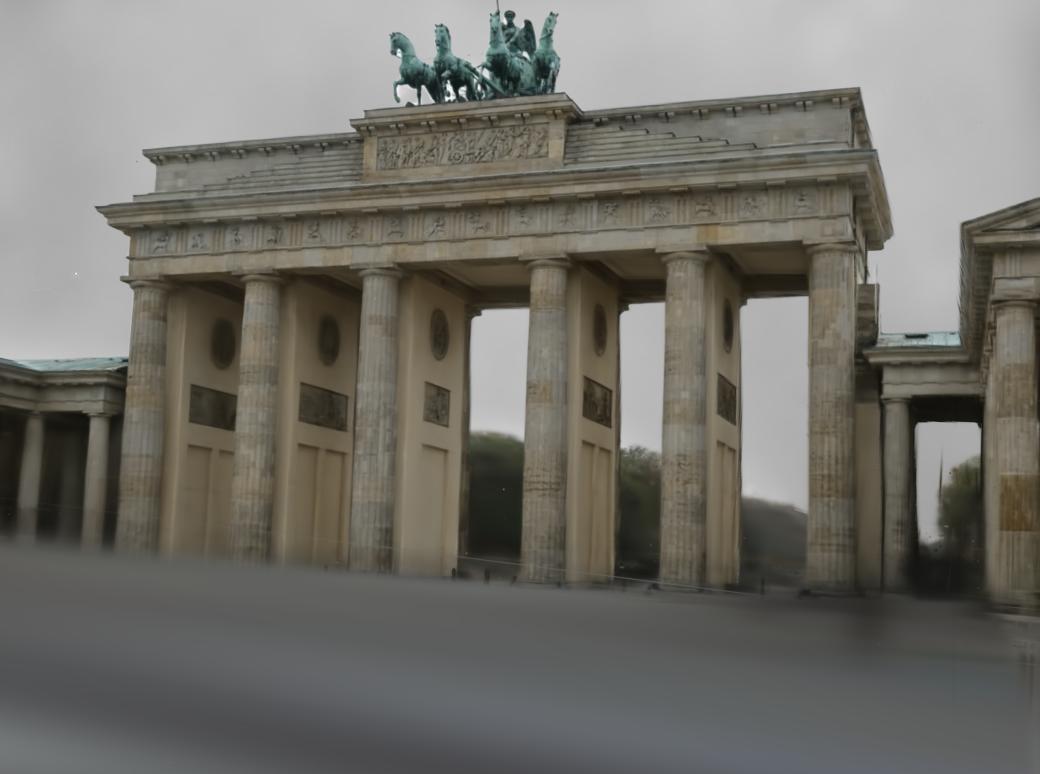}}
	\\
	\subfloat[\label{fig:e}]{
		\includegraphics[scale=0.1]{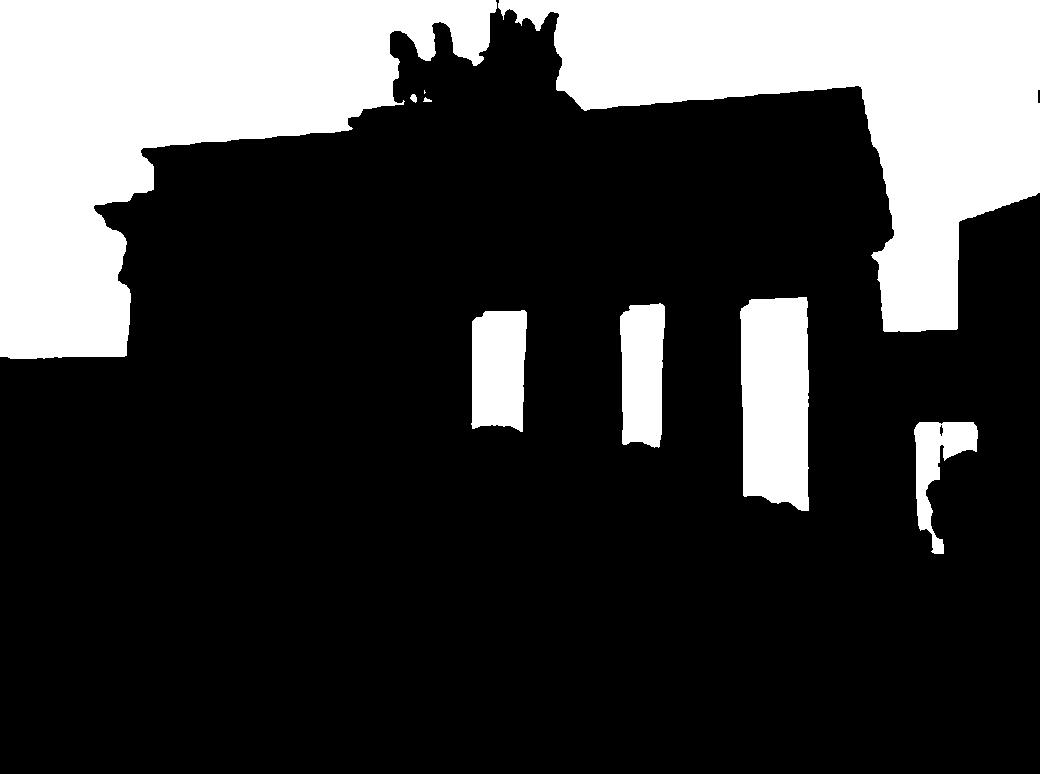}}
	\subfloat[\label{fig:f}]{
		\includegraphics[scale=0.1]{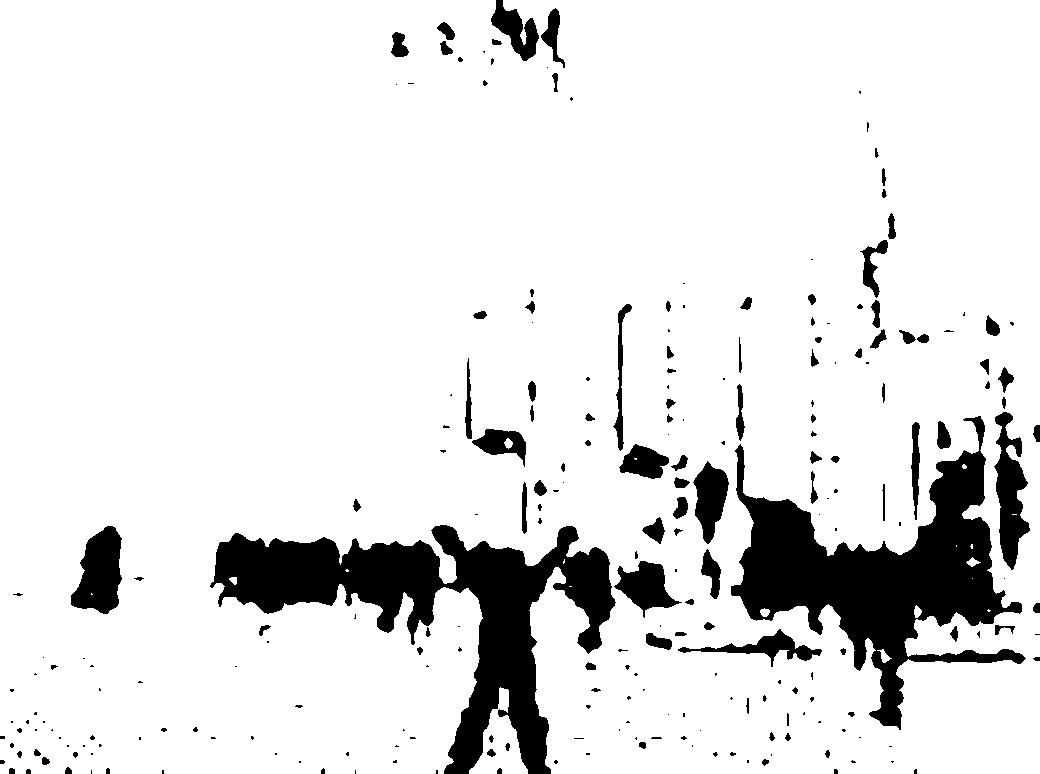}}
  	\subfloat[\label{fig:g}]{
		\includegraphics[scale=0.1]{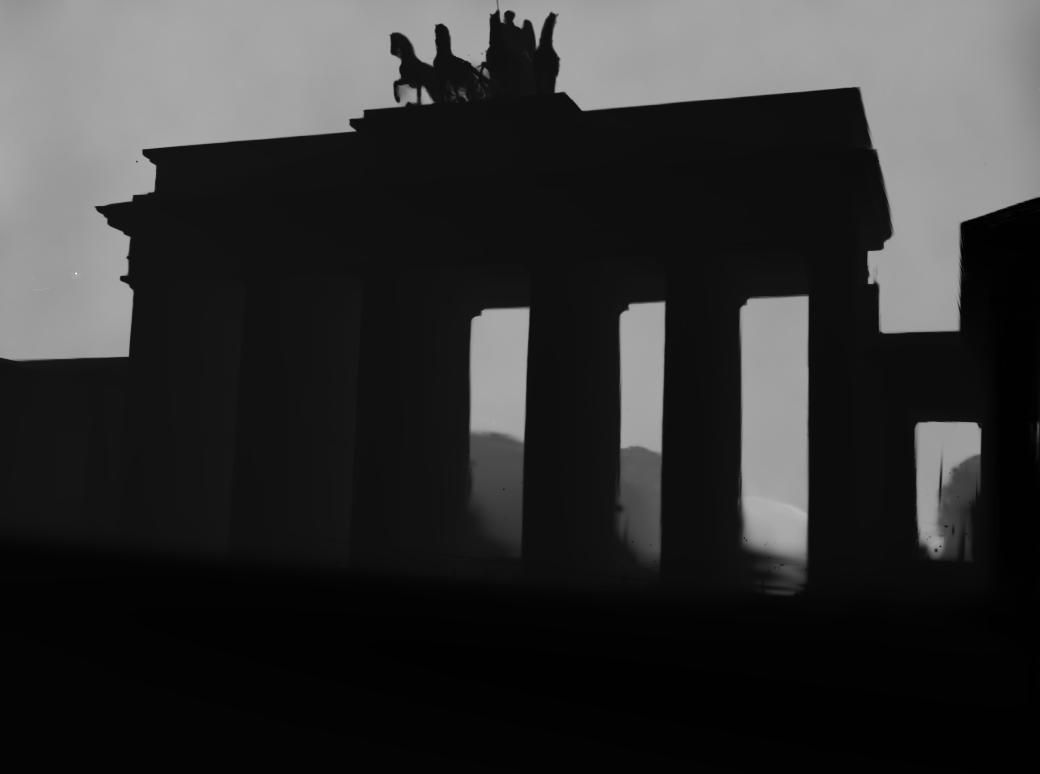}}
	\subfloat[\label{fig:h}]{
		\includegraphics[scale=0.1]{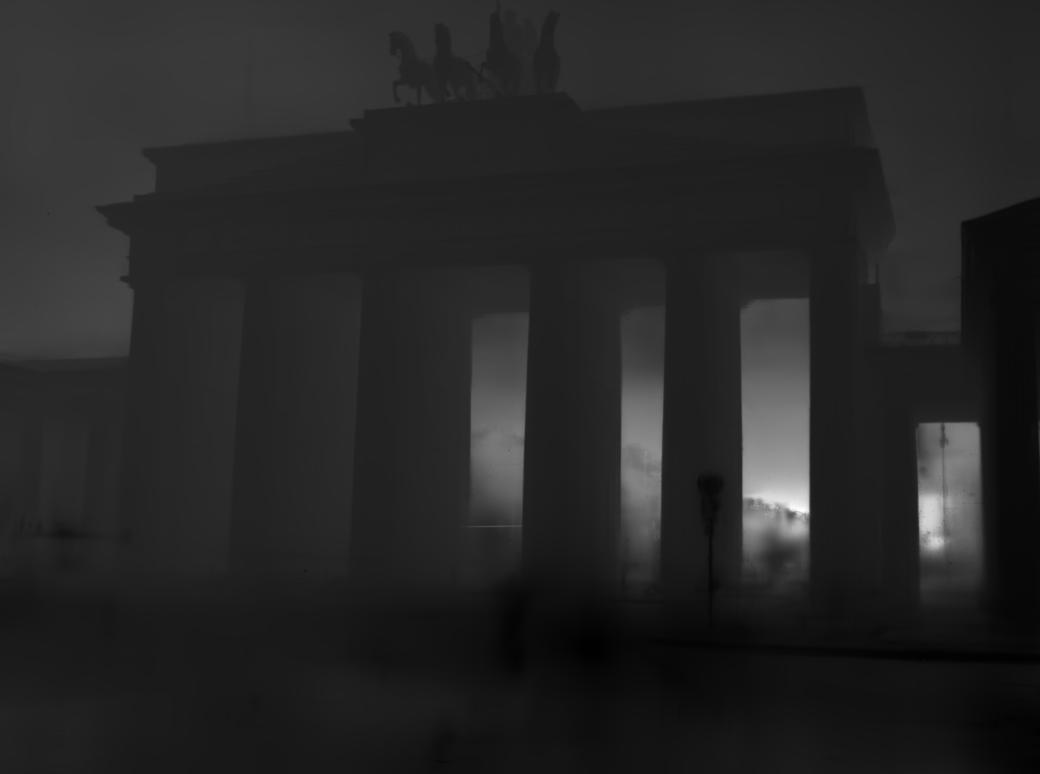}}
	\caption{Ablation study results: (a) Ground truth; (b) rendered result; (c) rendered result without the visibility map; (d) rendered result without sky handling; (f) visibility map; (g) rendered depth; (h) rendered depth without sky handling. Notably, in the absence of the visibility map, the rendered result exhibits more artifacts on the floor. Additionally, without sky handling, the depth of the sky overlaps with the foreground, leading to inconsistencies.}
	\label{fig:ablation} 
\end{figure*}

In concurrent works, the sky handling is often overlooked, which results
in the reconstruction of the infinitely distant sky at positions very close to the viewpoint, leading to artifacts as shown in Fig.\ref{fig:ablation}. These artifacts can significantly affect the viewing experience. In our work. We divide the foreground and sky apart during training. During the initialization phase, we follow the method outlined in paper \cite{kulhanek2024wildgaussians}, creating a sufficiently large sphere that encompasses the entire scene. We then use the Fibonacci sphere sampling algorithm \cite{10.1145/2816795.2818131} to sample 20,000 points on the sphere, which are used as Gaussian to represent the sky. We set the scene radius $r_s$ to 97\% of the $L_2$ norms of the centered input 3D points, while the sky sphere radius is defined as 10 times $r_s$.
OneFormer \cite{jain2023oneformer} is utilized generate sky masks. In the training process, the foreground and sky are separated and optimized individually, one is called foreground-Gaussians and the other is called sky-Gaussians. Foreground-Gaussians are responsible for rendering the foreground region, while sky-Gaussians handle the rendering of the sky. To ensure that the sky remains aligned with the sphere, the positions ($xyz$) of the sky Gaussians are fixed and do not participate in the updates. To prevent the foreground Gaussians responsible for rendering the foreground region from expanding into the background during the Gaussian densification process, we employed a gaussian rasterizer capable of forward rendering and backward gradient propagation with respect to both alpha and depth. The following losses were applied to enforce constraints. 
\begin{figure*}[!htbp]
	\centering
	\includegraphics[width=0.8\textwidth]{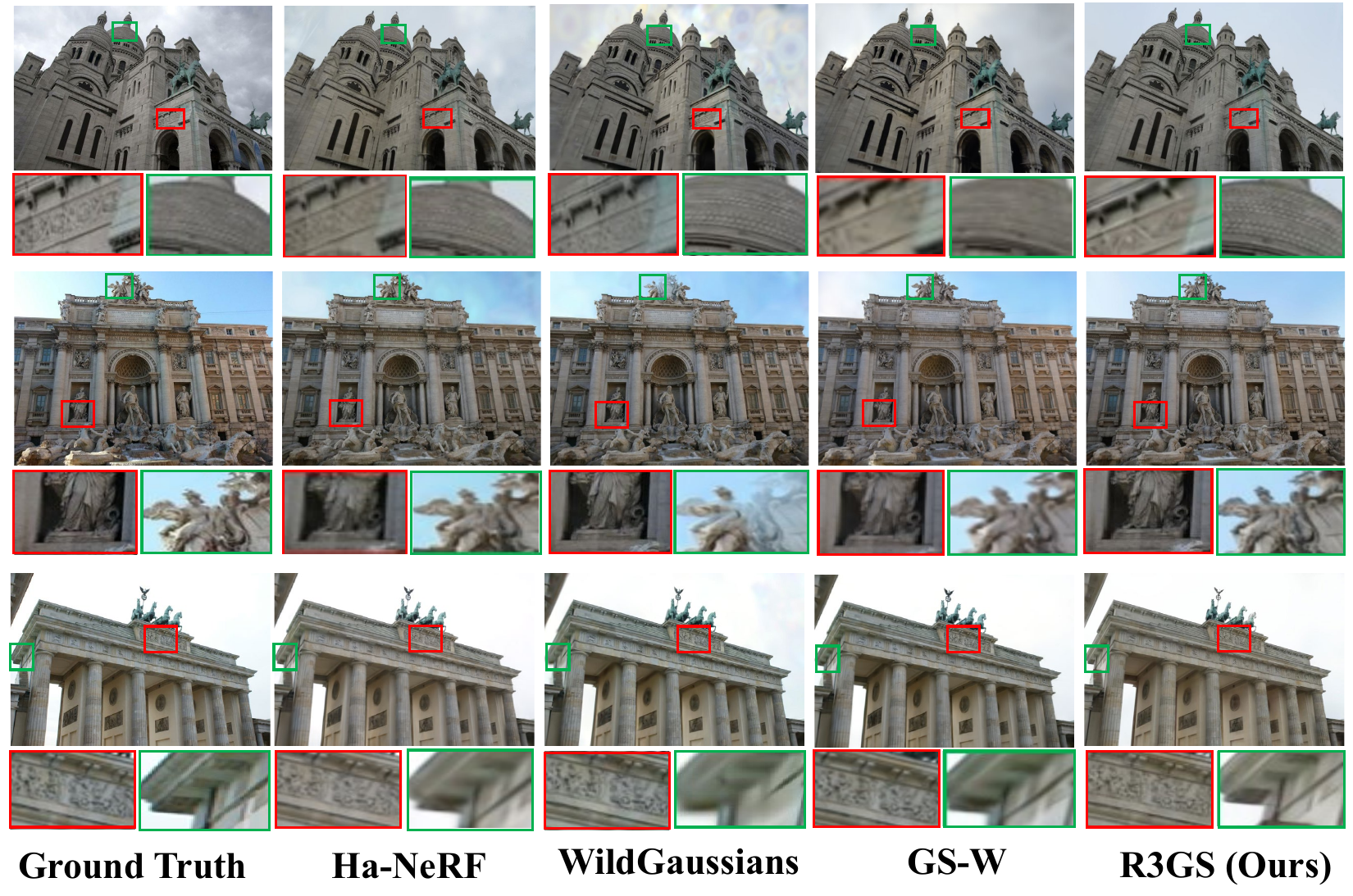}
	\caption{Visual comparison of rendering quality between different approaches. Red and green crops emphasize geometry differences. }
	\label{fig:comparison}
\end{figure*}

\subsection{Relocalization in outdoor scenes}
Most current 3DGS-SLAM pipelines are primarily applied to indoor scenes, as indoor datasets typically exhibit more consistent lighting conditions, which better satisfy the photometric consistency assumptions required by direct methods However, in real outdoor scenes, such as the NeRF-in-the-wild dataset, the photometric consistency assumption is difficult to satisfy for several reasons: 1) The varying appearance in the dataset. 2) Interference from transient objects like pedestrians and vehicles. 3) The continuously changing sky. 

In this section, a novel relocalization pipeline is proposed, which better satisfies the photometric consistency assumption, thereby achieving more robust relocalization performance. As shown in Fig. \ref{fig:pipeline}, given a query image, the trained CNN network is utilized to extract its appearance feature. Then, the appearance feature is firstly concatenated with the per-gaussian feature queried from Hash Grid and then fed into the MLPs to export standard 3DGS model. 

For the given image $I_{ref}$, we employ CosPlace \cite{Berton_2022_CVPR}, a place recognition system, to extract its feature representation $feat_{ref}$. This feature is then compared with a feature bank of all training images to retrieve the 3 nearest neighbors by computing the $L_{1}$ distance between features. The corresponding images of these nearest neighbors are considered to be the most geographically relevant and their rotations ${q_1, q_2, q_3}$ and translations ${t_1, t_2, t_3}$ and $L_{1}$ distance ${dist_1, dist_2, dist_3}$ are obtained. The init pose for the given image by using equation (\ref{eq:init_pose1}) and equation (\ref{eq:init_pose2}). 
\begin{align}
w_{i} &= \frac{1/dist_i}{\sum_{i=1}^3 (1/dist_i)}\\
q_{init} &= \sum_{i=1}^3 w_{i}q_{i} \label{eq:init_pose1} \\
t_{init} &= \sum_{i=1}^3 w_{i}t_{i} \label{eq:init_pose2} 
\end{align}
where $w_{i}$ is the weight and images with closer feature distances contribute more to the calculation of the initial pose $\boldsymbol{T}_{init}$. $\boldsymbol{T}_{init}$ is the init transformation matrix calculated from $q_{init}$ and $t_{init}$.
Following the 3DGS SLAM work\cite{matsuki2024gaussian,keetha2024splatam, Matsuki:Murai:etal:CVPR2024}, we use photometric loss $L_{p h o}$ to guide the pose optimization.
\begin{equation}
L_{p h o}=\left\|I\left(\mathcal{G}, \boldsymbol{T}_{C W}\right)-I_{ref}\right\|_1
\end{equation}
where $I\left(\mathcal{G}, \boldsymbol{T}_{C W}\right)$ renders the Gaussians $\mathcal{G}$ from $\boldsymbol{T}_{C W}$, and $I_{ref}$ is the given image. In the optimization process, the Gaussians $\mathcal{G}$ is fixed and only the pose $\boldsymbol{T}_{C W}$ is optimized. The gradient computation is same as MonoGS \cite{Matsuki:Murai:etal:CVPR2024}.

\section{EXPERIMENTS}
\subsection{Experimental Setup}
\textbf{Implementation Details.} All experiments were run on a single NVIDIA GeForce RTX 4090. Our model utilizes an anchor-based strategy, inspired by Scaffold-GS \cite{lu2024scaffold}, where $n_{offsets}$ is set to 5, resulting in each anchor generating 5 neural Gaussians. For appearance embedding, we employ a standard CNN to extract a 64-dimensional feature representing per-image appearance. Each anchor also queries the Multiresolution Hash Grid for a 32-dimensional per-anchor feature. These concatenated features are input into three shallow MLPs to predict opacity, color, and covariance. The network is optimized using the Adam algorithm \cite{kingma2014adam}. For the visibility map, we begin with a lightweight pre-trained model trained on the COCO dataset \cite{puri2019coco}. This model is capable of generating human masks, and after 5000 iterations of training, it demonstrates the ability to converge on other transient objects. In sky handling part, 20,000 initial anchors are placed in the sky sphere with a radius 10 times the radius of the initial point cloud(3\% outliers are excluded). 

\textbf{Datasets and Metrics.} Our experiments are conducted on Phototourism datasets \cite{snavely2006photo}, which including three scenes: Brandenburg Gate, Trevi Fountain, and Sacre Coeur. All of them include changing appearances, transient objects and different sky. Similarly, we split the training and test sets following the approach outlined in \cite{martin2021nerf}.

For evaluation, we assess performance on the test datasets using three widely adopted metrics: peak signal-to-noise ratio (PSNR), structural similarity index (SSIM) \cite{wang2004image}, and learned perceptual image patch similarity (LPIPS) \cite{zhang2018unreasonable}. In the relocalization phase, two metrics are used to evaluate the accuracy of the relocalization: Absolute Trajectory Error ($ATE$) and Rotation Distance ($R_{dist}$). 

\begin{align}
ATE&=( \left\|\operatorname{trans_i}-trans_{gt}||^2\right)^{\frac{1}{2}} \\
R_{\text {dist }}&= \arccos \left(\frac{\operatorname{tr}\left(\mathbf{R}_{\mathrm{i}} \mathbf{R}_{\mathrm{gt}}^{\mathrm{~T}}\right)-1}{2}\right) 
\end{align}

where $\mathbf{R}_{g t}$ and $\mathbf{R}_{i}$ represent the ground truth rotation matrix and the optimized rotation matrix, respectively. And $tr$ is the trace of a matrix. To quantify the translation error, we use the distance between the ground truth translation vector $trans_{gt}$ and $trans_i$.
\subsection{Comparison with State-of-the-art Methods} \textbf{Novel View Synthesis.} As illustrated in Fig. \ref{fig:comparison} and TABLE \ref{tab:1}, our approach delivers state-of-the-art performance in the NVS task. While directly optimizing 3DGS on in-the-wild photo collections results in subpar outcomes, NeRF-W and Ha-NeRF demonstrate moderate enhancements through the use of per-image features. CR-NeRF performs competitively using a cross-ray approach. Our method outperforms all concurrent works in PSNR, SSIM, and LPIPS. GS-W shows the best performance, but its results are based on downsampled images. Using the original resolution would result in memory overflow.

\textbf{Speed and Memory.} As shown in TABLE \ref{tab:1}, R3GS achieves the fastest training and rendering speeds (excluding GS-W, which downsamples images). It allows extracting appearance from any image and exporting it as a standard 3DGS point cloud, maintaining rendering speeds comparable to original 3DGS. Unlike other 3DGS methods, R3GS is memory-efficient, avoiding high-dimensional features for each Gaussian. Instead, it stores anchor features in a compact hash grid and prefilters visible anchors for each image during training, further improving efficiency. The use of a streamlined visibility map also contributes to faster training.

\textbf{Relocalization.} In the relocalization experiments, we perform relocalization for all images in the test set, using the poses from COLMAP \cite{fisher2021colmap} in the dataset as ground truth. Each relocalization process optimizes the pose for 500 iterations using Adam as the optimizer. The optimized poses are then compared with the ground truth poses by calculating the ATE and Rotation Distance metrics. For the Trevi Fountain, Sacre Coeur, and Brandenburg Gate datasets, we exclude the highest and lowest values in each group and calculate the average from the remaining data. The results are presented in TABLE \ref{tab:3}. We also integrated the same relocalization pipeline into the 3DGS system as a comparative experiment. The results demonstrate that R3GS achieves superior relocalization accuracy. This improvement is attributed to R3GS's ability to extract appearance from the current image and generate a 3DGS model that more effectively satisfies the photometric consistency assumption required by direct methods. 
\begin{table}[!htbp]\centering
	\centering
	\caption{Relocalization results on Phototourism datasets \cite{snavely2006photo}}\label{tab:relocalization}.
	\begin{tabular}{lllll}
		\toprule
		\multirow{2}{*}{\textbf{Scene}} & \multicolumn{2}{c}{\textbf{3DGS}} & \multicolumn{2}{c}{\textbf{R3GS}} \\
		&  ATE↓ & Rot Dist↓ & ATE↓ & Rot Dist↓  \\
		\midrule
		  Brandenburg Gate           &  0.401 & \textbf{0.576} & \textbf{0.101} & 0.582   \\ 
		Sacre Coeur       	       &  0.381 & 0.618 & \textbf{0.021} & \textbf{0.561}  \\
		Trevi Fountain      	   &  1.377 & 0.604 & \textbf{0.367} & \textbf{0.457}   \\
		\bottomrule
	\end{tabular}
\end{table}
\vspace{-1em}
\subsection{Ablation Study}
\begin{table}[t!bp]
\caption{Ablation studies on Brandenburg Gate.}
\scriptsize
\centering
\begin{tabular}{lccccc} 
\toprule
\multicolumn{1}{l}{\multirow{2}{*}{Method}} & \multicolumn{5}{c}{Brandenburg Gate}  \\ 
\cmidrule(lr){2-6}
\multicolumn{1}{c}{}                        & PSNR↑  & SSIM↑   & LPIPS↓  & Training↓ & FPS↑    \\ 
\hline
w/o sky                       & 26.51  & 0.9421    & 0.1327   & 0.303   & \textbf{241} \\
w/o vis map                   & 25.96  & 0.9226    & 0.1576   & \textbf{0.281}    & 225  \\ 
Ours full                     & \textbf{27.02}  & \textbf{0.9240}    & \textbf{0.1317}   & 0.311    & 230  \\
\bottomrule
\multicolumn{5}{l}{Training↓ means the time (hours) costing by 10000 iterations.}\\
\multicolumn{5}{l}{Rendering FPS↑ are calculated without any caching.}\\
\end{tabular}
\label{tab:3}
\end{table}

\textbf{Without visibility map.} Removing the transient object handling module can accelerate both training and rendering; however, the presence of transient objects introduces numerous artifacts in the rendered images. It's important to note that the testing data lacks transient objects, so the evaluation metrics on the testing set do not exhibit a significant drop.

\textbf{Without sky handling.} The sky handling module is pre-generated using the pre-trained OneFormer model and stored in the cache, so removing it has little impact on training speed. However, without the sky processing module, the point cloud file lacks the sky's point cloud, causing sky changes to converge more slowly. Additionally, we found that color changes in the sky lead the visibility map to treat the sky as a transient object, which significantly disrupts the learning of the sky's appearance.

\section{Conclusions}
In this paper, we introduce R3GS, a method capable of reconstruction and relocalization from unconstrained image collections. This method significantly enhances 3DGS's ability to synthesize novel views in unconstrained image collections. By utilizing appearance embedding with global features and hash coding, pre-trained visibility maps, and a sky handling mechanism, R3GS achieves improvements in performance metrics such as PSNR, SSIM, and LPIPS across several challenging datasets compared to GS-W, WildGaussians, and produce more detailed rendering effects while maintaining real-time rendering capabilities. Additionally, we propose a novel relocalization method based on our R3GS, which shows significant improvements in unconstrained image collections. Despite these advancements, certain challenges remain, particularly in rendering fine details under complex appearance variations and capturing high-frequency sky details. Upcoming research will concentrate on employing more complex network structures and special Gaussian properties to represent complex appearance and sky details.

\bibliographystyle{ieeetr}
\bibliography{bib/ref} 

\end{document}